\documentclass{article}

\usepackage{arxiv}

\usepackage[utf8]{inputenc} 
\usepackage[T1]{fontenc}    
\usepackage{hyperref}       
\usepackage{url}            
\usepackage{booktabs}       
\usepackage{amsfonts}       
\usepackage{nicefrac}       
\usepackage{microtype}      
\usepackage{lipsum}         
\usepackage{fancyvrb}       
\usepackage{comment}        
\usepackage{wrapfig}        
\usepackage{cleveref}       
\usepackage[nottoc]{tocbibind}  
\usepackage[export]{adjustbox}  

\usepackage{todonotes}

\newcommand{\version}{1.0}

\title{The Design Of ``Stratega'': \\ A General Strategy Games Framework}

\author{
  Diego Perez-Liebana\\
  Game AI Research Group\\
  Queen Mary University London\\
  \texttt{diego.perez@qmul.ac.uk} \\
  \And
  Alexander Dockhorn \\
  Game AI Research Group\\
  Queen Mary University London\\
  \texttt{a.dockhorn@qmul.ac.uk} \\
   \And
  Jorge Hurtado Grueso\\
  Game AI Research Group\\
  Queen Mary University London\\
  \texttt{j.hurtado@qmul.ac.uk}
   \And
  Dominik Jeurissen \\
  Game AI Research Group\\
  Queen Mary University London\\
  \texttt{djeurissen.paper@web.de}
}

\begin{document}
\maketitle

\begin{abstract}
Stratega, a general strategy games framework, has been designed to foster research on computational intelligence for strategy games.
In contrast to other strategy game frameworks, Stratega allows to create a wide variety of turn-based and real-time strategy games using a common API for agent development.
While the current version supports the development of turn-based strategy games and agents, we will add support for real-time strategy games in future updates.
Flexibility is achieved by utilising YAML-files to configure tiles, units, actions, and levels.
Therefore, the user can design and run a variety of games to test developed agents without specifically adjusting it to the game being generated.
The framework has been built with a focus of statistical forward planning (SFP) agents.
For this purpose, agents can access and modify game-states and use the forward model to simulate the outcome of their actions.
While SFP agents have shown great flexibility in general game-playing, their performance is limited in case of complex state and action-spaces.
Finally, we hope that the development of this framework and its respective agents helps to better understand the complex decision-making process in strategy games.
Stratega can be downloaded at:

\centering\url{https://github.research.its.qmul.ac.uk/eecsgameai/Stratega}
\end{abstract}

\rule{\textwidth}{2pt}

\begin{center}
\includegraphics[width=0.985\textwidth, fbox]{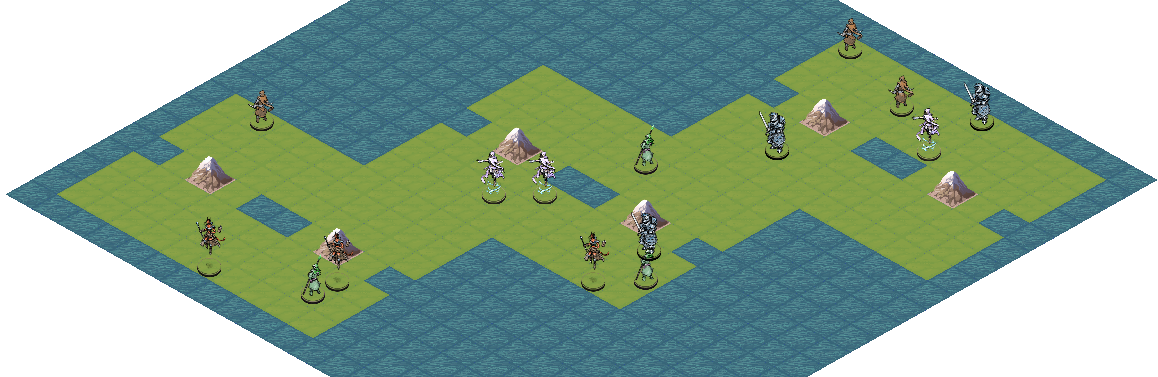}
\end{center}

\newpage

\rule{\textwidth}{2pt}
\tableofcontents
\rule{\textwidth}{2pt}


\section{How to use this Document?}

This paper serves the purpose of a living documentation for Stratega - a general strategy games framework for studying computational intelligence in turn-based and real-time strategy games. 
We aim to continuously adjust the framework to the needs of our users.
This paper will be updated accordingly and describe the inner workings of the Stratega framework as well as the underlying design philosophy.

When citing this paper please make sure to specify its version since the paper will be updated over time!

\section{The Stratega Framework}

Due to their high complexity, strategy games have become an interesting genre for AI research~\cite{buro2003real}.
Fundamental characteristics can include, but are not limited to: complex state and action spaces, partial observability, spatial and temporal reasoning, competition and collaboration in multiple-player settings, player modelling, among others.

Existing strategy game frameworks, e.g. Starcraft~\cite{bwapi, vinyals2017starcraft, ontanon2013survey} and $\mu$RTS~\cite{ontanon2018first}, focus on studying a single game.
This not just allows to incorporate domain knowledge during AI development but also indefinitely training the agent to become proficient in this game.
It has been shown that this approach can result in strong AI agents capable of defeating even the best human players, e.g. the AlphaStar agent for Starcraft II~\cite{alphastarblog}

Due to these recent successes in development of single strategy game AI, we believe the time is right to propose a general strategy games framework.
Using a similar approach as general game-playing frameworks such as GVGAI~\cite{perez2019general} and GGP~\cite{genesereth2005general}, the Stratega framework allows to create a variety of strategy games using a common API for AI development.
Stratega allows the creation of n-player turn-based and real-time strategy games with full access to the game's forward model.
While other general game frameworks may be able to generate similar games, the strong focus of Stratega allows to incorporate higher level concepts which ease the development process.
In this paper, we are outlining the design of the Stratega framework and discuss its future prospects.

\section{Design and Overview of the Framework}
\label{sec:overview}
\begin{figure*}[t]
    \centering
    \includegraphics[width=\textwidth, trim=1cm 0 0 0, clip]{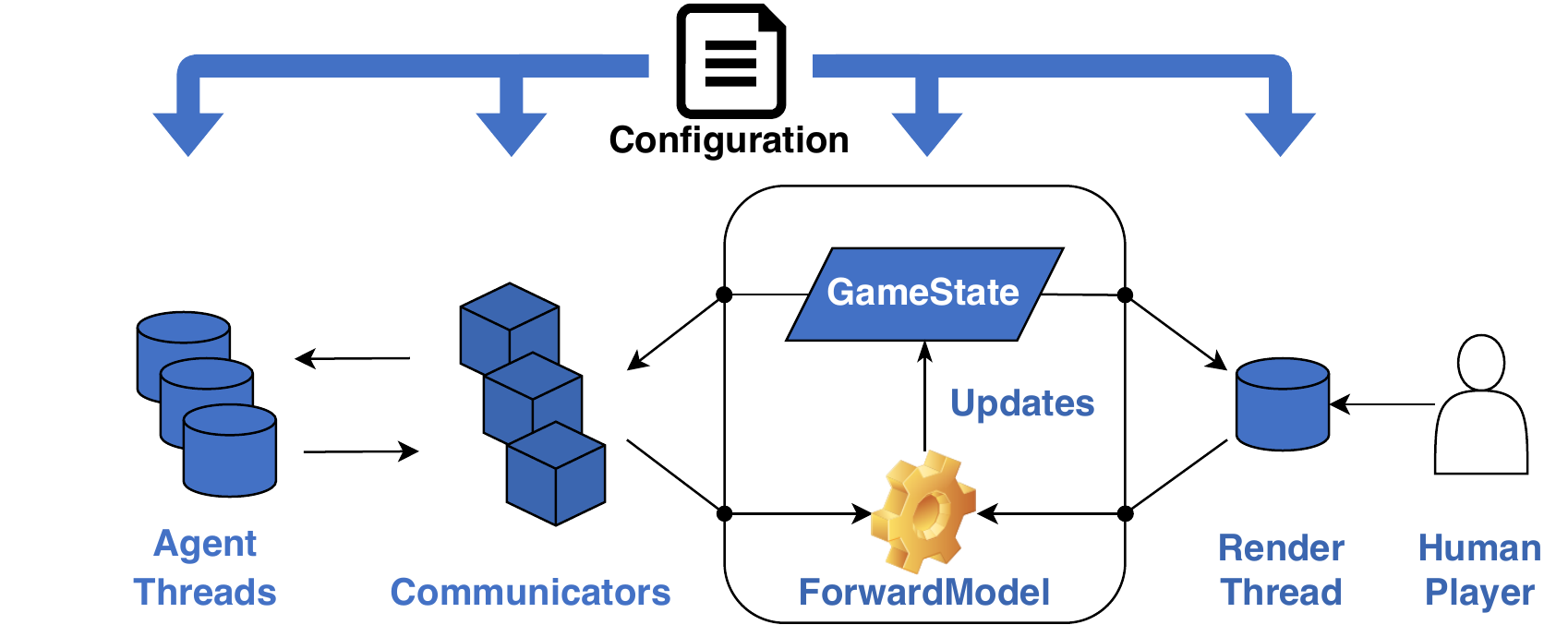}
    \caption{Schematic overview of the framework's architecture.}
    \label{fig:my_label}
    \vspace{1cm}
\end{figure*}

The Stratega framework mainly consists of 4 parts, i.e. the game-runner, rendering, agents, and configuration.

The core of the framework is the game-runner, whose sole purpose is to run games and provide information about the game. The framework provides two game-runners, one for turn-based and one for RTS-games.
The agents are responsible for controlling the actions of a player in the game state. 
To do this, each agent receives a game communicator to send actions or poll information about the current game state. Additionally, each agent has access to a copy of the forward model of the game-runner.
Regardless of the type of game, every agent-thread runs until the game is over. Agents can query information or send actions at any point in time, even during an opponent's turn. This enables agents to think outside their own turns. Additionally, the agent can observe how the game evolves to think ahead. Note that the turn-based game-runner ignores actions from players when it's not their turn. Meanwhile, the RTS-runner will collect the actions from all players and execute them in order of receipt after a set amount of time.

The runner keeps track of which agent controls which player in the game state. If no agent is available for a player, the runner will send a signal and wait until an external source provides the missing actions. 
The render-thread subscribes to events from the game-runner to update the window. If the runner signals missing agents, the GUI will switch to interactive mode to let human-players play the game. That also has the added benefit that human-players have no time limit.
This approach has similarities to MVVM, in the sense that the game-runner represents the view-model. The renderer (view) is decoupled from the logic and can be easily disabled to switch between headless-execution and GUI-execution.

A feature of the framework is the ability to write algorithms for various degrees of game-abstractions. While game-states represent nodes in a game-tree, the forward model represents the edges to get from one node to another. That means the forward model provides actions, as well as the game-logic.
Developers can provide a new forward model for the default game state or provide their new representations. Algorithms implemented in the framework can work with any kind of forward model.

Lastly, the framework provides the ability to configure games via YAML-files, which are explained in more detail in upcoming sections. It's noteworthy that configuration is separated from the rest of the framework, meaning the configuration initializes and sets up the game, but the framework does not need to be re-compiled every time a change in the configuration is made. 

\newpage
\section{Game-State}

The Game-state object serves as a data container without any logic.
It offers access to the current board, its tiles, a list of players, and their units.
The board is represented as a $n \times m$-grid of tiles.
A tile can be occupied by a unit and will therefore be blocked for other units.
Tile types and board layouts are defined via the configuration file (see \Cref{sec:configuration}).
A unit is a player controllable character which is able to perform one or multiple actions per turn.
Each player controls all its units at once, meaning that during a single turn the agent may return multiple actions until ending its turn.

Due to the strict separation of data and game logic, players receiving a game-state object can freely copy or modify it.
In case the game has been configured to be partially observable, the player's observation will be limited to tiles which are in line of sight of at least one of its units.
Information outside the range will be replaced by a default tile.
To ease the development of agents in partial-observable environment, we provide a method for completing a partially observable game-state.
For each unobserved tile, the method randomly samples the tile type and its contained units. 
The chances of sampling a unit can be influenced by the agent to introduce a pessimistic or optimistic bias.


\vspace{-0.25cm}

\section{Forward Models and Actions}
\label{sec:forwarad-model}

The forward model is the driving component of this framework.
Being provided with a game-state it will generate an ActionSpace object.
It handles the efficient creation of available actions and updates the set of actions after the game-state has been updated.
Returned action objects store the type of action, the executing unit and the target tile.
Nevertheless, they do not include the action's logic which remains part of the forward model.

When a player provides the forward model with an action and a game-state, it first checks if the action is applicable.
This needs to be done, since the player may have modified either the action or the game-state in between generating the set of actions and simulating its outcome.
These security checks can be deactivated to increase the framework's speed. 
In case the action can be executed, the game-state object will be advanced accordingly.

Implemented actions include a movement and an attack action as well as special abilities such as healing units among others.
The way in which actions are to be executed is defined in the forward model, which is set up using the configuration files.
This allows to quickly change the action's parameters, and therefore, create variants of the same game.
This approach has been chosen to easily modify the game's balance without updating the code.

Additional flexibility is achieved by the use of triggers and effects.
Given an event, e.g. a unit entering a cell or a player ending a turn, the forward model can execute subscribed methods.
This can be used to implement game-mechanics such as a unit losing health at the end of each turn.
Alternatively, it allows to log information every time an event occurs.

Both, the forward model and the game state, can be extended by the user to create game-state abstractions.
This can be useful for reducing the complexity of the game-states encoding and ease the training of AI agents.
In the future, we will implement a vector-based state observation to better support the development of reinforcement learning agents.
Similarly, search-based agents may achieve higher performance due to the reduced search space of such an abstraction.

\vspace{-0.25cm}

\section{Agents and Algorithms}
\label{sec:algorithm-and-agents}

The framework provides access to a small set of standard agents which will be extended over time.
Provided rule-based agents focus on different aspects of the game, e.g. combat, the use of special abilities, or exploiting their surroundings.
These agents serve as a baseline for evaluating the performance of more sophisticated agents and represent a great introduction into developing agents for our framework.

Additionally, we provide several scoring functions which can be used in conjunction with search-based agents.
With the first version of the framework, we provide basic implementations of classic search algorithms, such as  One-step look-ahead, Monte Carlo method, Monte Carlo Tree Search~\cite{Browne2012} (MCTS) as well as Rolling Horizon Evolutionary Algorithm~\cite{perez2013rolling} (RHEA).
While MCTS and RHEA have shown to perform well in general game-playing tasks~\cite{perez2019general}, they seem to struggle with the higher complexity of strategy games.
Initial experiments have shown that the performance of both search-based agents is still limited in comparison to the sample rule-based agents~\cite{strategaAIIDE20}, which motivates further research on extending their performance.

\section{Configuration}
\label{sec:configuration}

\begin{wrapfigure}[50]{r}{0.42\textwidth}
\vspace{-1.3em}
\hfill\begin{minipage}[t]{0.40\textwidth}
    \begin{Verbatim}[frame=single,fontsize=\footnotesize]
Tiles:
    Swamp:
        Symbol: S
        IsWalkable: true
    Mountain:
        Symbol: M
        IsWalkable: false
    Hole:
        Symbol: H
        IsWalkable: true
        
Board:
    GenerationType: Manual
    Layout: >
        MMMMM
        MSSSM
        MSSHM
        MSSHM
        MSSSM
        MMMMM
        
Units:
    Warrior:
        Health: 100
        MovementRange: 3
        LineOfSightRange: 4
        AttackDamage: 20
        Actions: [Attack, Move]
    Healer:
        Health: 40
        MovementRange: 5
        LineOfSightRange: 4
        HealAmount: 10
        Actions: [Heal, Move]
        
ForwardModel:
    WinCondition: LastManStanding
    Effects:
        DamageAll:
            Type: Damage
            Trigger: EndOfTurn
            Condition: None
            Amount: 10
        DeadlyHole:
            Type: Death
            Trigger: EnterTile
            Condition: StandingOnTile
            TargetTile: Hole
    \end{Verbatim}
    \caption{An exemplary configuration}
    \label{fig:configuration}
    \end{minipage}
\end{wrapfigure}

In this section, we will have a closer look on the YAML configuration file.
It allows users to adjust the balance of an existing game or define new game modes by simply modifying the game-components through keywords.
An exemplary configuration file is shown in \Cref{fig:configuration}.

First, we will take a loot at the definition of tiles and board.
Similar to the level description in the video game definition language~\cite{Schaul2014}, we first define tiles and their encoding in the tilemap.
Each tile definition consists of a symbol and if it is walkable or not.
Further tile attributes will be added during in upcoming versions of the framework.

Later on, the board can be manually described in terms of a tile-map, an $n$-by-$m$ grid of tiles.
Alternatively, it will be possible to procedurally generate levels.
Therefore, the user needs to provide a method which will receive the tile configuration and a set of parameters to generate a level.
This allows the framework to be used for research on level generation.

Next, units are defined in terms of unit types and their properties.
Each unit exhibits four basic attributes, i.e. health, attack damage, movement range, and line of sight range.
The health defines the number of damage a unit can take until it dies.
Losing health is not permanent and it can be replenished using abilities or other effects.
A unit's attack damage describes the number of damage a unit can deal with an attack.
Its movement range describes the number of tiles a unit can walk when choosing the Move action.
The line of sight is only considered in partial-observable games.
Here, each unit can observe tiles with a Manhattan distance equal or less than the given threshold.
A player's view of the board will be limited by the union of all tiles visible by their units.
Finally, each unit receives a list of actions.

On the right, we show the definition of a warrior and a healer.
While the warrior has high health and high attack damage, its movement range is very limited.
Since its only actions are attack and move, the warrior can be considered a strong melee unit. 
In comparison, the healer is a much weaker unit which is unable to attack.
Instead, it can use the action heal which replenishes the health of a damaged unit.

The last part of our configuration consists of the forward model's properties.
Here, we first defined a win condition.
The \textit{LastManStanding} option lets players fight until all but one have lost all their units.
With the \textit{Effects} keyword users can introduce more general game mechanics.
An effect's trigger describes when the effect will be executed.
Conditions can further limit the effects to be only applied to conforming units or tiles.
Here, the \textit{DamageAll} effect reduces each unit's health by 10 at the end of a turn.
In comparison, the \textit{DeadlyHole} effect triggers in case a unit steps on a hole, which will result in the unit being removed from the game.

In upcoming versions of our framework, we will extend the set of available keywords to ease the development of diverse game-modes.
We further look into the parameterisation of actions and tiles to increase the expressiveness of available keywords.

\section{Future Directions}
\label{sec:future-directions}
    
Our aim is to provide a framework for search-based and reinforcement learning agents in complex environments.
Additionally, we believe that the way in which game variants can be configured make the framework suitable for research in game and level generation, and automatic game tuning/balancing.

In its current state the framework fully supports tactical turn-based games and offers an extensive API for developing AI agents.
The framework's inherent graphical user interface and logging capabilities support the user to develop agents and analyse their game-playing capabilities.
Further logging capabilities will be included in future updates to facilitate research into explainability of AI agents and their decision-making process in complex environments.

In the future, we plan to extend the framework by many facets of modern turn-based and real-time strategy games.
Hence, our road-map includes the incorporation of tactical role-playing elements, technology and cultural trees as well as resource and economy management.
While this list will be updated over time, these components would allow to model a large variety of strategy games.
Furthermore, we will add full support of real-time strategy games to the framework.


\section*{Acknowledgements}
This work is supported by UK EPSRC research grant EP/T008962/1.

\bibliographystyle{unsrt}  
\bibliography{references}

\end{document}